# Curiosity-based Robot Navigation under Uncertainty in Crowded Environments

Kuanqi Cai, Weinan Chen, Chaoqun Wang, Hong Zhang, *Fellow, IEEE*, and Max Q.-H. Meng, *Fellow, IEEE*

*Abstract*—Mobile robots have become more and more popular in large-scale and crowded environments, such as airports, shopping malls, etc. However, due to sparse landmarks and crowd noise, localization in this environment is a great challenge. Furthermore, it is unreliable for the robot to navigate safely in crowds while considering human comfort. Thus, how to navigate safely with localization precision in that environment is a critical problem. To solve this problem, we proposed a curiosity-based framework that can find an effective path with the consideration of human comfort and crowds, localization uncertainty, and the cost-to-go to the target. Three parts are involved in the proposed framework: the distance assessment module, the Curiosity for Positive Content (CPC), namely information-rich areas, and the Curiosity for Negative Content (CNC), namely crowded areas. CPC is introduced when the real-time localization uncertainty evaluation is not satisfied. This factor is predicted through the propagation of uncertainty along the candidate trajectory to provoke the robot to approach localization-referenced landmarks. The Human Comfort and Crowd Density Map (HCCDM) based on the Gaussian Mixture Model (GMM) is established to calculate CNC, which drives the robot to bypass the crowd and consider human comfort. The evaluation is conducted in a series of large-scale and crowded environments. The results show that our method can find a feasible path that can consider the localization uncertainty while simultaneously avoiding the crowded area.

*Index Terms*—Motion and Path planning, service robotics, localization, autonomous vehicle navigation.

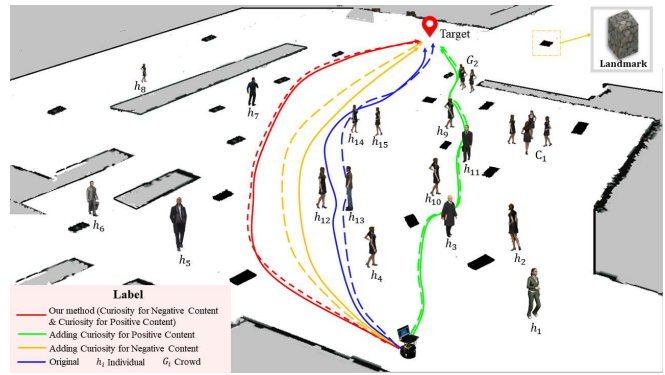

Figure 1. Illustration of the proposed method. Our method with Curiosity for Positive Content (information-rich areas), and Curiosity for Negative Content (crowded areas), is able to generate a human-friendly path (red) that can simultaneously avoid crowds and approach landmarks, which makes humans comfortable and minimizes state estimation uncertainty. Besides, the method with Curiosity for Positive Content (green) can generate a trajectory that approaches the landmark area, but it may lead to a high collision risk with humans. Additionally, the method of only considering Curiosity for Negative Content (yellow) does not perform well in localization. The traditional path planner (original) enters both crowded and landmark-deficient areas, resulting in large localization uncertainty and comfort reduction.

## I. INTRODUCTION

IN the past few years, with the rapid development of robots, service robots driving in terminals, shopping malls, exhibition malls, and so on have attracted growing attention [1]. The operating scope of service robots is relatively large in such environments that have dense crowds and lack landmarks.

Manuscript received July 17, 2022; Revised November 5, 2022; Accepted December 20, 2022.This paper was recommended for publication by Editor Aniket Bera upon evaluation of the Associate Editor and Reviewers' comments. This project is supported by the Shenzhen Key Laboratory of Robotics Perception and Intelligence ZDSYS20200810171800001, the National Natural Science Foundation of China 62103179. *(Corresponding author: Weinan Chen, Max Q.-H. Meng.)*

Kuanqi Cai and Hong Zhang are with the Department of Electronic and Electrical Engineering, Southern University of Science and Technology, Shenzhen, China. kayle.ckq@gmail.com; zhangh33@sustech.edu.cn

Weinan Chen is with School of Electro-mechanical Engineering, Guangdong University of Technology, China, weinanchen1991@gmail.com.

Chaoqun Wang is with the School of control science and Engineering, Shandong University, Shandong, China, chaoqunwang@sdu.edu.cn.

Max Q.-H. Meng is with Shenzhen Key Laboratory of Robotics Perception and Intelligence, and Department of Electronic and Electrical Engineering, Southern University of Science and Technology, Shenzhen 518055, China, on leave from the Department of Electronic Engineering, The Chinese University of Hong Kong, Hong Kong, and also with the Shenzhen Research Institute of The Chinese University of Hong Kong, Shenzhen 518057, China, max.meng@ieee.org

Consequently, state estimation is inaccurate because of the sparse landmarks and measurement noise generated by humans [2]. Besides, robots driving into crowded areas may cause the "freezing robot problem" that the robot either makes no forward progress or takes extreme evasive action to avoid collisions within the crowd, let alone consider human comfort [3]. Therefore, how to generate feasible trajectories in such environments becomes an essential problem for robots [4].

To solve the above problem, we proposed a curiosity-based planner focusing on human comfort, crowds, and localization uncertainty. From a psychological point of view, curiosity is the internal psychology of animals that instinctively want to add information about something when all or part of the information is blank. This information can be divided into Curiosity for Positive Content (CPC) and Curiosity for Negative Content (CNC). In our curiosity-based framework, curiosity is defined as the unsupervised act affected by the areas that contain different types of content based on the robots' current condition. We defined information-rich areas, which contribute to localization as CPC and defined crowds as CNC. In addition, we use the distance between humans and robots to quantify human comfort. Our proposed method aims to seek a feasible path considering both low localization uncertainty and human-comfort behavior to satisfy navigation in large and crowded environments. Therefore, when the positioning information of the robot is insufficient, CPC will be generated to inspire the robot to move to places with more landmarks to add information. Otherwise, CNC will update during the navigation process to provide crowd information, driving the robot away from the crowds.

The rest of the paper is outlined in the following ways. The related work is presented in Section II. Section III gives the statements of the problem formulation. Section IV shows the methodology. A series of experiments illustrate the efficiency of our method in comparison to prior research in Section V. Finally, the discussion and conclusion of the research study are presented in Section VI.

## II. RELATED WORK

How to realize the harmonious coexistence of humans and robots in the environment has been extensively studied over the last decade. In this section, we intersect two main topics of socially aware navigation systems represented by two main questions: (i) *"How to navigate in the crowded environment?"*, and (ii) *"How to mitigate localization uncertainty during navigation in large scale and crowded environments?"*.

Many studies to answer the question *"How to navigate in the crowded environment?"* can be found in the following literature. Some path planners have been successfully deployed in crowded environments and have achieved considerable success. Everett *et al.* [5] proposed a navigation scheme which is based on deep $RL$ and $LSTM$. Although this method can travel autonomously in crowded environments by following specific behavioral guidelines, it can not solve the "freezing robot" problem when robots move in a crowd. For mitigating the freezing problem, Cai *et al.* [6] developed a novel path planner that contains the Improved Virtual Doppler Method ($IVDM$) and the Collision Risk and Human Space ($CR\&HS$) modules, which can drive robots to bypass the crowd. Aiswarya *et al.* [7] used the RRT algorithm for path planning and collision avoidance in a collaborative 2D workspace. Narayanan *et al.* [8] developed an unified method for pedestrian-aware navigation, which has high efficiency in crowded environments. Majd *et al.* [9] combined time-based RRTs with Control Barrier Functions (CBFs) to develop safe mobility plans in dynamic situations with a large number of pedestrians. However, they cannot guarantee stable high-speed operation and meet the requirements of human comfort in large, crowded environments due to the lack of localization uncertainty consideration.

Several robot navigation frameworks have been implemented and demonstrated to answer the question *"How to mitigate the localization uncertainty during navigation in large scale and crowded environments?"*. Zhang *et al.* [10] suggested using a receding horizon strategy to create a trajectory that takes into account localization uncertainty, collision risk, and path length. However, the computation cost rises as a result of the expansion of the planning space to include all restrictions. Tyler [11] introduced the distributionally robust RRT (DR-RRT), which considers moment-based ambiguity sets of distributions incorporating localization uncertainty, and external environmental uncertainty. To balance localization uncertainty and energy efficiency, an optimum trajectory planning technique based on an upgraded dolphin swarm algorithm is suggested in [12]. Pepy *et al.* [13] presented an efficient way to plan paths in an uncertain-configuration space using Safe-RRT, which can find safe paths despite the robot uncertainty. Similar approaches have also been published in [14], [15] to address this issue. Although the above tactics can help robots localize accurately, they fall short in extremely dynamic environments. As can be seen from the analysis above, the dense crowd and sparse landmarks in the large environment increase the observation noise of the robot. Thus, taking human awareness and localization uncertainty into account at the same time during the navigation process is still unsolved.

Therefore, to propose a feasible solution to the above problem, we proposed a path planner to facilitate human-friendly path planning while decreasing localization uncertainty in large-scale and crowded environments. For accurate localization, a real-time uncertainty evaluation strategy and CPC are introduced to reduce the localization uncertainty. Besides, the Human Comfort and Crowd Density Map (HCCDM) is proposed to calculate CNC for the robot to generate a human-friendly motion plan in cluttered areas.

## III. PROBLEM FORMULATION

As mentioned above, the key point of navigation in a large-scale and crowded environment is to find a feasible path to bypass the crowd and reduce the uncertainty of localization. We designed a curiosity-based cost function and incorporated it into the rapidly exploring random tree-base path planner scheme to find a feasible trajectory. We use $\mathbb{M}$ to represent the map of the environment created by the robot. $\mathbb{O}_{obs}(t)$ represents obstacles including humans vector $ped(t)$ and landmarks vector $\mathbb{O}_{mark}$. $\mathbb{O}_{free}(t)$ represents the free space at time t. The robot's systems are nonlinear and only partially observable. The discrete-time description of its dynamics and sensors is depicted as:

$$x_t = f(x_{t-1}, u_{t-1}, m_t), \quad m_t \sim \mathcal{N}(\mathbf{0}, M_t) \quad (1)$$
$$z_t = g(x_t, n_t), \quad n_t \sim \mathcal{N}(\mathbf{0}, N_t) \quad (2)$$
$$x_t = [x'', y'', theta], \quad z_t = [z_t^1, ..., z_t^K]. \quad (3)$$

$x_t \in \chi$ is the valid robot state vector in the state space $\chi$. $u_t \in U$ is the control vector in the control space $U$. $m_t$ ($n_t$) is the noise with mean 0 and variance $M_t$ ($N_t$). $x''$, and $y''$ are two-dimensional plane coordinates of the robot, and $theta$ is the angle. $z_t^k$ is the independently measured distance. The measurement vector of the robot is $z_t \in Z$ and $Z$ is the observation space.

During the navigation process, the path planning is repeated at each time step $\Delta t$. $q_j^i : \{q_j^1, q_j^2, ..., q_j^I\}$ represents nontrivial candidate trajectories generated in $j_{th}$ time step. $q_{j,K}^i : \{[x_{j,1}^i, u_{j,1}^i]^T, ..., [x_{j,K}^i, u_{j,K}^i]^T\}$ contains a number of states, control inputs. $k$ is the index of state along the path. The optimal path from a set of nontrivial trajectories in $j_{th}$ time step can be formulated by

$$\begin{aligned} \mathcal{Q}_{opt} = & \min \mathcal{L}'(q_j^i) \\ \text{s.t.} & \quad q_{j,1}^i = f_{root}, \\ & \quad q_j^i \in \mathbb{O}_{free}(t), \quad \forall t \in [t, t + \Delta t]. \end{aligned} \quad (4)$$

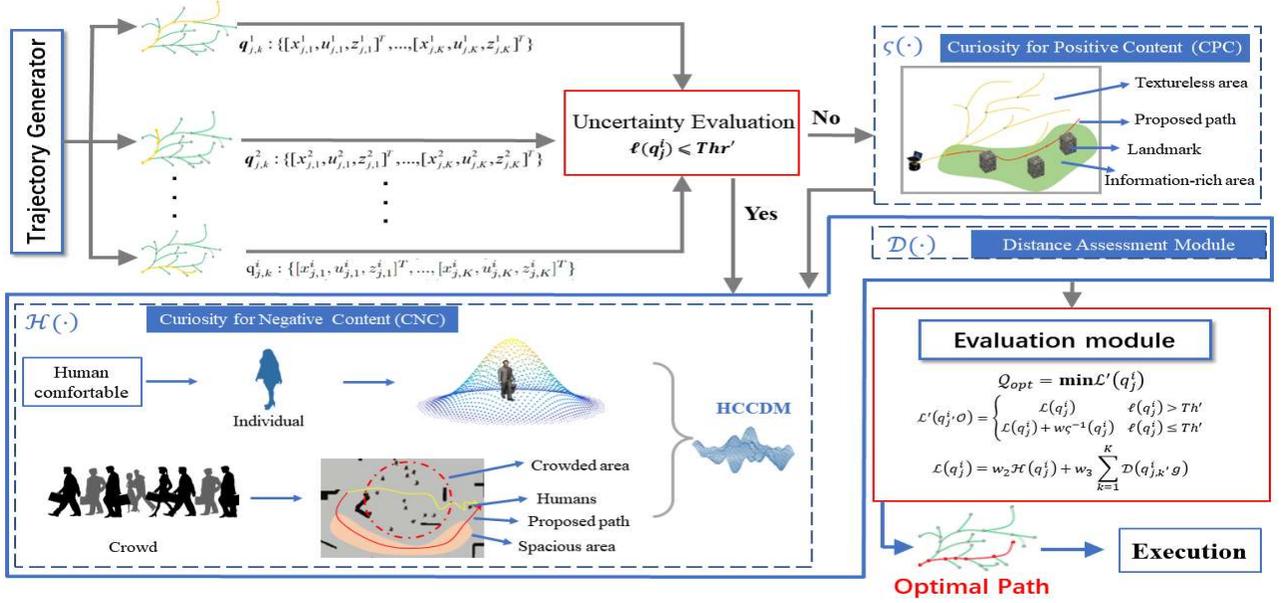

Figure 2. The curiosity-based path planner's system diagram. **Evaluation Module** computes the cost of the various trajectories created by the **Trajectory Generator**, and the trajectory with the lowest cost is deemed the best.

$q_{j,1}^i$ is the starting point of the robot at $j_{th}$ time step and $f_{root}$ is the root of the rapidly-exploring random tree. $\mathcal{L}'(\cdot)$ is the objective function to find the optimal one from a set of nontrivial trajectories and is expressed as

$$\mathcal{L}'(q_j^i) = \begin{cases} \mathcal{L}(q_j^i) & \ell(q_j^i) \leq Thr' \\ \mathcal{L}(q_j^i) + w_1\varsigma^{-1}(q_j^i) & \ell(q_j^i) > Thr'. \end{cases} \quad (5)$$

$\ell(\cdot)$ is the evaluation of localization uncertainty and $Thr'$ is the localization threshold. When $\ell(q_j^i)$ is higher than a given threshold, we consider localization to have failed. $\varsigma(\cdot)$ represents the formulation of CPC. This means that when the localization is uncertain, CPC will yield to navigating the robot to the information-rich area. $w_1$ is the weight. $\mathcal{L}(\cdot)$ is the socially aware cost function. It consists of the distance assessment module and CNC. CNC will increase, when the robot moves into a crowded area. Therefore, the robot will be drawn to an open area to avoid crowds. Moreover, considering both CPC and CNC, the robot is attracted to landmark-rich areas with few pedestrians. The formula of $\mathcal{L}(\cdot)$ is:

$$\mathcal{L}(q_j^i) = w_2 \mathcal{H}(q_j^i) + w_3 \sum_{k=1}^{K} \mathcal{D}(q_{j,k}^i, g), \quad (6)$$

where $\mathcal{D}(\cdot)$ is the distance assessment module, which is similar to the traditional method [16]. $\mathcal{H}(\cdot)$ represents CNC containing human comfort and human density. The weight $w_1$, $w_2$, and $w_3$ are optimized manually according to a series of offline tests. When $w_1$ increases, the robot will drive close to the area with more landmarks. As $w_2$ increases, the robot will be far away from the crowd. Adjusting $w_3$ makes the robot choose a trajectory with a short distance to run. In sum, the target of this paper is to find $q_j^i$ that minimizes the cost function in equation 4.

**Algorithm 1:** Trajectory Generation

**Input:** Map $\mathbb{M}$, Goal $g$
**Output:** Trajectory candidates $q_j^i$

1  $q_j^i = \emptyset$;
2  $Tree \leftarrow InitializeTree()$;
3  $q_{rand} \leftarrow Sampling(\mathbb{M})$;
4  $q_{near} \leftarrow Nearest(q_{rand}, Tree)$;
5  $[q_{new}, \tau] \leftarrow Steer(q_{rand}, q_{new})$;
6  $Tree = Extend(q_{near}, q_{rand})$;
7  **if** $ObstacleFree(q_{new})$ **then**
8  $\quad q_{neighbor} \leftarrow FindNearNeighbor(Tree, q_{new})$;
9  $\quad q_{min} \leftarrow Parent(q_{neighbor}, q_{near}, q_{new})$;
10 $\quad Tree.add(q_{min}, q_{new}, \tau)$;
11 $\quad Tree \leftarrow Rewire(q_{neighbor}, q_{new}, Tree)$;
12 $q_j^i \leftarrow FindPathCandidates(Tree)$;

## IV. METHODOLOGY

The system for trajectory creation and assessment in the planning stage is shown in Fig. 2. The rapidly exploring random tree-based planner scheme is used in the **Trajectory Generator**, which constructs a list of potential paths from the robot's present position to the next position. Second, we calculate the localization uncertainty of the current robot position. If the localization uncertainty is higher than the threshold, CPC will be introduced in the **Evaluation Module** and the trajectory with lower localization uncertainty will be rewarded and vice versa. Third, the **Evaluation Module** is leveraged for the best path with minimum cost.

### A. Framework Design

The workflow of **Trajectory Generator** can be seen in Alg. 1. $Sampling(\cdot)$ is used to generate the random point $q_{rand}$

**Algorithm 2:** Optimal Trajectory Generation

**Input:** Trajectory candidates $q_j^i$, Current state $x_{t-1}$
**Output:** Optimal path from a set of candidates $\mathcal{Q}_{opt}$

1 Humans $ped$, Landmark $\mathbb{O}_{mark}$;
2 **while** $g$ not reached **do**
3     Observe($\mathbb{O}_{mark}$, $\mathbb{O}_{obs}$);
4     Delete unreachable trajectories($q_j^i$, $x_c$, $t$);
5     Predict $ped$ at time $t,...,t+N*\Delta t$;
6     $\ell = UncertaintyCalculate(x_{t-1})$;
7     **if** $\ell > Thr'$ **then**
8        Introduce CPC $P_{CPC}$;
9     **else**
10        $P_{CPC}=0$;
11     Build K-D Tree $T'(ped)$;
12     **for** $i <$ Number of candidates **do**
13        **for** $k < size(q_j^i)$ **do**
14           Distance Assessment Set $P_{Dis}$;
15           Introduce CNC Set $P_{CNC}$;
16           $k$++;
17        $i$++;
18     $\mathcal{L}' = CostCalculate(P_{Dis}, P_{CNC}, P_{CPC})$;
19     $\mathcal{Q}_{opt} \leftarrow \text{Min}(\mathcal{L}')$;
20     **if** $\mathcal{Q}_{opt} = \emptyset$ **then**
21        break;
22     **else**
23        move along $q_{opt}$ for one step;

TABLE I
PARAMETERS AND DESCRIPTION

| Symbol | Description |
| --- | --- |
| $\Sigma_t$ | State estimation covariance matrix |
| $\overline{\Sigma}_t$ | Covariance matrix of prediction |
| $M_t'$ | Variance of state transition uncertainty |
| $\Lambda_t$ | Uncertainty of not having yet taken observations |
| $L_t$; $K_t$ | Kalman gain; State feedback gain |
| $\mu_t$ | Mean state estimation covariance matrix |

in $\mathbb{O}_{free}$. $Nearest(\cdot)$ is to search $Tree$ for the nearest point $q_{near}$ to $q_{rand}$. $Steer(\cdot)$ extends $Tree$ from $q_{near}$ to $q_{rand}$ with path $\tau$ considering the kinematic constraint of robots. $q_{new}$ represents the end of the path $\tau$. $q_{near}$ is the neighbor point of $q_{new}$. $FindNearNeighbor(\cdot)$ is the function to reselect the neighbor point of $q_{new}$ on $Tree$. $Rewire(\cdot)$ is the rewiring process of $Tree$ to reduce redundant length. These processes are repeated and the $Tree$ continuously updates during each time step $\Delta t$. Candidates on the $Tree$ for robot navigation are generated through $FindPathCandidates(\cdot)$. $FindPathCandidates(\cdot)$ is the function to select the trajectories that are collision-free and conform to the robot motion model as candidates.

The procedure of effective path generation based on **Evaluation Module** is shown in Alg. 2. During the navigation process, the robot updates its observations (see lines 3 to 5). Moreover, **Evaluation Module**, which contains CNC, CPC, and distance assessment modules, is executed to find an optimal trajectory from a set of candidates (see lines 6 to 23).

### B. Curiosity for Positive Content (CPC)

*1) Evaluation of Localization Uncertainty:* We defined CPC as a probabilistic model that is designed to reduce the localization uncertainty of the robot in large-scale and crowded environments. We construct the following time-varying linear system by taking the appropriate partial derivatives of equations (1) and (2).

$$x_t' = A_t x_{t-1}' + B_t u_{t-1}' + m_t', \quad m_t' \sim \mathcal{N}(\mathbf{0}, M_t') \quad (7)$$

$$z_t' = C_t x_t' + n_t', \quad n_t' \sim \mathcal{N}(\mathbf{0}, N_t'). \quad (8)$$

$x_t'$, $u_{t-1}'$, and $z_t'$ are error quantities. Intuitively, $x_t = x_t^* + x_t'$, $x_t^*$ is the deterministic nominal value. However, $x_t$ will not be accessible throughout execution. Therefore, $\hat{x}_t$ is defined to estimate the true value $x_t$, $x_t \sim \mathcal{N}(\hat{x}_t, \Sigma_t)$. To determine whether CPC should be considered, we measure the uncertainty of the current state $\mathbf{x}_i$ by calculating the trace of the state estimation covariance matrix $\Sigma_t$. The uncertainty of the robot's current position $\ell$ can be measured quantitatively as:

$$\ell(x_{t-1}) = Trace(\Sigma_{t-1}). \quad (9)$$

*2) Calculation of CPC:* In the situation of large localization uncertainty, we calculated CPC by uncertainty prediction [2] to drive the robot to an area with high localization accuracy. We use a Kalman filter to maintain the Gaussian state estimate. A process step predicting the associated covariance of the next state is:

$$\overline{\Sigma}_t = A_t \Sigma_{t-1} A_t^T + M_t' \quad (10)$$

$$\Sigma_t = \overline{\Sigma}_t - L_t C_t \overline{\Sigma}_t, \quad (11)$$

where $L_t$ is the Kalman gain. The recursive function for the mean is shown below:

$$\mu_t = (A_t - B_t K_t)\mu_{t-1}. \quad (12)$$

$K_t$ is the state feedback gain. Moreover, the distribution of the state estimates is:

$$P(x_t) = N(x_t^*, \Lambda_t + \Sigma_t) \quad (13)$$

$$\Lambda_t = (A_t - B_t K_t)\Lambda_{t-1}(A_t - B_t K_t)^T + L_t C_t \overline{\Sigma}_t. \quad (14)$$

$P(x_t)$ defines the distribution over trajectories as the sum of the online state estimate covariance plus the uncertainty caused by the lack of observation value. The uncertainty of one nontrivial trajectory $q_j^i$ is

$$\zeta(q_j^i) = \sum_{k=1}^{K} Trace(\Lambda_{j,k}^i + \Sigma_{j,k}^i). \quad (15)$$

The higher the uncertainty of position is, the less curious it is to the robot. The definitions of the variable are discredited in Table I.

## C. Curiosity for Negative content (CNC)

We first establish the Human Comfort and Crowd Density Map (HCCDM), which is based on GMM to calculate CNC. The establishment of HCCDM is divided into three steps: the crowded cluster, HCCDM construction, and HCCDM update.

*1) Online Crowd Cluster:* To quickly classify crowds and individuals online to satisfy real-time planning, a KD-tree is set up with individuals $ped_i = (x_i, y_i)$ as input. Then, we do a range search on the established tree with social distance[1] as a threshold. When the distance between individuals is less than the social distance, we assume these individuals form a crowd.

*2) HCCDM Construction:* After clustering, a mapping process based on GMM is performed to represent the crowd and human comfort. To the best of our knowledge, the Gaussian Model (GM) is the most efficient algorithm to model the human comfort of individuals, which is time-consuming in a crowded environment. Therefore, we only build the human density model of crowds instead of individuals. Bypassing the crowd can ensure that the individuals' comfort in the crowd is not affected. In individual terms, human comfort will be considered. These two models are based on the two-dimensional Gaussian Model (GM). The probability density function is defined as:

$$\mathcal{N}(q|\mu_g, \Sigma_g) = \frac{1}{2\pi|\Sigma_g|^{1/2}} \exp\{-\frac{1}{2}(q-\mu_g)^T \Sigma_g^{-1}(q-\mu_g)\}, \quad (16)$$

where $\mu_g$ is the mean, and $\Sigma_g = \begin{bmatrix} \sigma_g & 0 \\ 0 & \sigma_g \end{bmatrix}$ is the covariance matrix. $\mu_g$ and $\sigma_g$ are given by the numbers of human ($\varepsilon$) in the cluster. The function is defined as:

$$\begin{cases} \mu_g = ped_i, & \sigma_g = D_{pd} & \varepsilon = 1 \\ \mu_g = \mathcal{C}(\boldsymbol{ped_g}), & \sigma_g = \mathcal{R}(\boldsymbol{ped_g}) + D_{pd} & \varepsilon > 1. \end{cases} \quad (17)$$

$D_{pd}$ is defined as the personal distance [17]. $\mathcal{C}(\cdot)$ is used to find the center of the circumcircle formed by pedestrians $\boldsymbol{ped_g} = \{ped_n, n = 1, 2, \cdots, \varepsilon\}$ in the cluster. The function $\mathcal{R}(\cdot)$ is defined to calculate the radius of the circumcircle. The function of HCCDM is shown below. $p(k')$ is the weight of the component of the probability density function. The CNC of $q_j^i$ in HCCDM is calculated by

$$\mathcal{H}(q_j^i) = \sum_{k'=1}^{K'} p(k')\mathcal{N}(q_{j,k'}^i|\mu_g, \Sigma_g). \quad (18)$$

*3) HCCDM Update:* Since not all pedestrians will affect robot navigation (e.g., pedestrians who are far away from the robot), Working Zone [6] is used to represent the work area of the robot. To reduce computation, we update HCCDM by filtering pedestrians leaving the Working Zone, because these pedestrians have no impact on the navigation of the robot.

## V. SIMULATIONS AND RESULTS

### A. Experiment Setup

We carry out experimental studies based on the Robot Operating System (ROS) and a computer with Ubuntu 16.04 is adopted as the simulation platform[2]. The robot in the simulation environment is mounted with a 3D laser sensor (Sick Laser). The Adaptive Monte Carlo Localization algorithm (AMCL) [3] is used for localization. In order to make the simulation close to the real environment, humans move in different directions and speeds within the environment. We designed four basic environments, which are shown in Fig. 3. Thirteen scenarios are generated by changing the number of humans, laser observation rates, and landmark numbers in four basic environments. Notably, the maps of scenarios after changing the number of landmarks are slightly different from those of base environments. The parameter configuration of different scenarios is shown in Table II. The configuration of different environments is designed based on the control variable method, and the specific value of the parameter is given randomly. In Env3 (S*), we use the Hybrid Reciprocal Velocity Obstacle (HRVO) [18], which is one of the popular methods to simulate human motion. The laser observation rate is defined by the ratio of the laser observation area to the map size. A lower laser observation rate indicates a larger environmental scale. These maps are built by Occupancy Grid Map[4]. The landmark number is the ratio of the number of landmarks occupied by grids to the total number of grids.

In terms of the comparison algorithm, to show the superiority of the proposed method, planners that consider CPC or CNC only, and two human-aware path planners (Risk-RRT [19], Bi-Risk-RRT [20]) are used in the comparison. The experiment is repeated ten times for each algorithm in each scenario. For a more comprehensive evaluation of the algorithm, the evaluation of planners is split into three parts: localization evaluation, human awareness evaluation, and efficiency evaluation.

### B. Localization Evaluation

From the perspective of the localization, we used the root mean square error ($RMSE$) [21] and the ratio of the localization covariance matrix's trace convergence time to running time ($TCM$) as metrics. $RMSE$ is used to evaluate the localization accuracy, and $TCM$ is used to assess the time of the first convergence of localization. Besides, the number of mutations ($NM$) represents the number of vibrations caused by inaccurate localization after the first convergence. Experimental results of different methods in twelve scenarios are shown in Table III.

As Table III shows, methods with curiosity for information-rich areas (our method and adding CPC) have the lowest $RSME$, $TCM$, and $NM$. Besides, with the decrease in lidar observation rates and landmark numbers, methods with the proposed CPC can still maintain the lowest values. It means that with a lower $RSME$ value, the proposed method achieves accurate localization during the navigation process. In addition, the proposed module can reach the state of accurate localization in a short time and can provide continuous accurate localization, which has a lower $TCM$ and $NM$.

---

[1] https://en.wikipedia.org/wiki/Proxemics
[2] Video demonstration available at https://youtu.be/t83CQcAm9O4
[3] http://wiki.ros.org/amcl
[4] https://en.wikipedia.org/wiki/Occupancy_grid_mappings

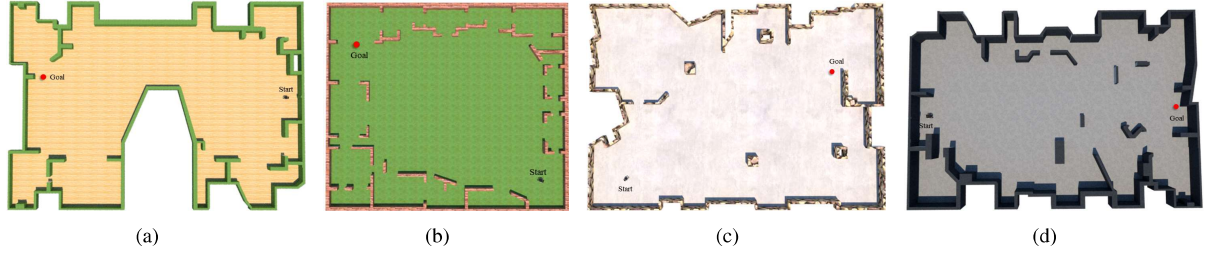

Figure 3. Qualitative testing of the planner in four different maps. (a) Environment 1. (b) Environment 2. (c) Environment 3. (d) Environment 4.

Table II
EXPERIMENTAL PARAMETER CONFIGURATION. THE LASER OBSERVATION RATE IS DEFINED BY THE RATIO OF THE LASER OBSERVATION AREA TO THE MAP AREA. LANDMARK'S NUMBER IS EQUAL TO THE RATIO OF THE NUMBER OF GRIDS OCCUPIED BY A LANDMARK TO THAT BY MAPS.

|  | Env1 | | | Env2 | | | Env3 | | | | Env4 | | |
|---|---|---|---|---|---|---|---|---|---|---|---|---|---|
|  | S1 | S2 | S3 | S1 | S2 | S3 | S1 | S* | S2 | S3 | S1 | S2 | S3 |
| Number of human | 25 | 31 | 37 | 28 | 34 | 40 | 25 | 25 | 30 | 40 | 27 | 27 | 40 |
| Ratio (human/m$^2$) | 25/378.30 | 31/363.82 | 37/363.82 | 28/272.21 | 34/272.21 | 40/272.21 | 25/433.91 | 25/433.91 | 30/433.91 | 40/433.91 | 27/420.14 | 27/420.14 | 40/420.14 |
| Laser observation rate | 4.35% | 4.52% | 4.52% | 8.87% | 3.94% | 3.94% | 7.43% | 7.43% | 3.79% | 3.79% | 7.67% | 3.91% | 3.91% |
| Landmark number | 28946/303983 | 31397/292351 | 31397/292351 | 16274/218736 | 16274/218736 | 20254/218737 | 17970/348668 | 17970/348668 | 17970/348668 | 19401/348668 | 26121/337606 | 30645/337606 | 26121/337606 |

Table III
LOCALIZATION METRICS

|  |  | Our method (CPC & CNC) | | | Adding CPC | | | Adding CNC | | | Risk-RRT | | | Bi-Risk-RRT | | |
|---|---|---|---|---|---|---|---|---|---|---|---|---|---|---|---|---|
|  |  | RMSE ↓ | TCM ↓ | NM ↓ | RMSE ↓ | TCM ↓ | NM ↓ | RMSE ↓ | TCM ↓ | NM ↓ | RMSE ↓ | TCM ↓ | NM ↓ | RMSE ↓ | TCM ↓ | NM ↓ |
| Env1 | S1 | **0.34**±0.01 | 5.82%±0.32 | **0** | 0.34±0.02 | **5.26%±0.20** | **0** | 0.46±0.12 | 36.74%±11.54 | 2 | 0.50±0.13 | 67.04%±29.08 | 1 | 0.38±0.06 | 22.16%±12.51 | 2 |
|  | S2 | **0.33**±0.01 | 6.38%±0.40 | **0** | **0.33±0.01** | **5.67%±0.39** | 1 | 0.38±0.05 | 35.48%±6.24 | 1 | 0.34±0.01 | 31.21%±5.24 | 1 | 0.36±0.02 | 31.80%±5.68 | 1 |
|  | S3 | **0.33**±0.01 | 6.34%±0.47 | **0** | **0.33±0.01** | **5.51%±0.41** | 1 | 0.37±0.03 | 28.21%±7.58 | 1 | 0.34±0.03 | 30.87%±1.71 | 1 | 0.37±0.03 | 23.38%±7.76 | 1 |
| Env2 | S1 | **0.15**±0.01 | 14.01%±0.35 | **0** | 0.18±0.02 | 15.37%±2.56 | **0** | **0.15±0.02** | 15.30%±1.91 | **0** | 0.18±0.02 | 20.52%±7.83 | 1 | 0.18±0.01 | 14.72%±1.94 | 1 |
|  | S2 | **0.15**±0.01 | **11.89%±0.5** | **0** | 0.16±0.01 | 13.05%±1.50 | 1 | 0.18±0.02 | 15.96%±2.49 | 2 | 0.26±0.04 | 21.96%±6.89 | 1 | 0.21±0.03 | 21.66%±12.70 | 1 |
|  | S3 | **0.14**±0.01 | **10.37%±0.4** | **0** | 0.15±0.15 | 11.30%±1.91 | **0** | 0.18±0.03 | 20.41%±4.54 | 1 | 0.20±0.04 | 14.57%±3.47 | 1 | 0.20±0.05 | 12.86%±1.68 | 1 |
| Env3 | S1 | **0.20**±0.01 | 6.51%±0.54 | **0** | **0.20±0.01** | **6.25%±0.35** | **0** | 0.23±0.04 | 6.57%±0.22 | 3 | 0.24±0.02 | 7.16%±0.21 | 2 | 0.27±0.04 | 6.79%±0.29 | 2 |
|  | S* | 0.22±0.02 | 7.13%±0.23 | **0** | 0.23±0.02 | **4.96%±0.09** | **0** | 0.25±0.03 | 9.01%±4.77 | 2 | 0.59±0.44 | 10.13%±2.97 | 2 | 0.37±0.25 | 13.13%±6.65 | 1 |
|  | S2 | **0.20**±0.02 | 7.10%±0.92 | **0** | 0.21±0.04 | **6.46%±0.24** | **0** | 0.24±0.04 | 7.74%±0.59 | 2 | 0.44±0.24 | 7.10%±0.44 | 1 | 0.30±0.05 | 34.10%±37.26 | 2 |
|  | S3 | 0.21±0.01 | 7.14%±0.27 | **0** | **0.21±0.01** | **6.86%±0.25** | 1 | 0.31±0.08 | 9.60%±+2.73 | 2 | 0.33±0.07 | 7.22%±0.90 | 2 | 0.32±0.07 | 27.69%±33.26 | 3 |
| Env4 | S1 | **0.29**±0.01 | 8.61%±1.27 | **0** | 0.30±0.01 | 8.77%±1.25 | **0** | 0.32±0.02 | 14.99%±7.24 | **0** | 0.37±0.03 | 25.78%±15.43 | **0** | 0.40±0.06 | 42.21%±3.94 | **0** |
|  | S2 | **0.29**±0.01 | **6.63%±0.18** | **0** | **0.29±0.01** | 6.86%±0.40 | **0** | 0.39±0.06 | 44.67%±8.72 | 1 | 0.48±0.06 | 51.25%±8.95 | 1 | 0.40±0.02 | 41.89%±3.63 | **0** |
|  | S3 | 0.31±0.02 | 7.10%±0.34 | **0** | 0.32±0.02 | **6.44%±0.36** | **0** | 0.33±0.03 | 8.20%±1.77 | 1 | 0.56±0.05 | 69.89%±2.26 | **0** | 0.57±0.14 | 69.86%±10.23 | **0** |

Table IV
HUMAN AWARENESS METRICS

|  |  | Our method (CPC & CNC) | | | Adding CPC | | | Adding CNC | | | Risk-RRT | | | Bi-Risk-RRT | | |
|---|---|---|---|---|---|---|---|---|---|---|---|---|---|---|---|---|
|  |  | TD ↓ | MD ↑ | NT ↓ | TD ↓ | MD ↑ | NT ↓ | TD ↓ | MD ↑ | NT ↓ | TD ↓ | MD ↑ | NT ↓ | TD ↓ | MD ↑ | NT ↓ |
| Env1 | S1 | **0** | **1.89±0.26** | **0** | 2.28%±2.25 | 1.40±0.22 | 1 | 6.78%±8.64 | 1.32±0.26 | 2 | 12.46%±8.64 | 0.95±0.49 | 5 | 18.56%±11.14 | 0.76±0.22 | 8 |
|  | S2 | **1.57%±1.55** | **1.51±0.25** | 1 | 3.42%±3.12 | 1.37±0.40 | **1** | 8.17%±3.02 | 1.07±0.22 | 2 | 18.05%±8.60 | 0.85±0.06 | 4 | 17.30%±5.19 | 0.86±0.19 | 3 |
|  | S3 | **1.72%±1.86** | **1.51±0.27** | 1 | 3.57%±3.72 | 1.30±0.33 | **1** | 7.41%±4.83 | 1.00±0.19 | 2 | 11.74%±4.37 | 0.91±0.21 | 4 | 20.52%±8.33 | 0.83±0.03 | 3 |
| Env2 | S1 | **0** | **1.70±0.04** | **0** | 23.93%±18.67 | 0.92±0.25 | 4 | 6.50%±2.50 | 0.94±0.12 | 2 | 28.06%±9.02 | 0.73±0.27 | 10 | 26.49%±6.40 | 0.83±0.13 | 5 |
|  | S2 | **0.93%±0.91** | **1.57±0.20** | 1 | 9.32%±8.23 | 1.07±0.37 | 4 | 11.69%±2.65 | 0.97±0.16 | 3 | 18.32%±14.49 | 0.95±0.12 | 5 | 20.09%±5.82 | 0.95±0.21 | 4 |
|  | S3 | **4.04%±1.20** | **1.16±0.19** | 1 | 5.10%±2.31 | 1.11±0.25 | 2 | 25.15%±12.67 | 0.95±0.11 | 4 | 40.60%±25.43 | 0.82±0.10 | 6 | 25.90%±5.34 | 0.93±0.17 | 5 |
| Env3 | S1 | **0.11%±0.24** | **2.10±0.47** | **0** | 5.72%±1.67 | 1.19±0.19 | 2 | 9.16%±6.53 | 0.95±0.10 | 16 | 26.54%±9.62 | 0.64±0.33 | 16 | 31.52%±21.54 | 0.96±0.12 | 3 |
|  | S* | **0** | **2.39±0.48** | **0** | 8.54%±5.76 | 0.94±0.22 | 2 | 12.92%±10.57 | 0.49±0.34 | 3 | 20.85%±15.70 | 0.76±0.26 | 3 | 18.13%±12.76 | 0.85±0.26 | 4 |
|  | S2 | **1.68%±1.86** | **1.48±1.86** | 1 | 4.80%±4.63 | 1.32±0.63 | **1** | 4.31%±1.70 | 1.04±0.17 | 2 | 22.84%±10.14 | 0.67±0.39 | 4 | 28.36%±16.89 | 0.87±0.12 | 5 |
|  | S3 | **1.22%±2.19** | **1.64±0.37** | 1 | 6.18%±4.66 | 1.10±0.30 | 2 | 16.84%±3.19 | 0.91±0.22 | 5 | 45.91%±14.67 | 0.78±0.10 | 7 | 36.39%±12.63 | 0.89±0.19 | 7 |
| Env4 | S1 | **0** | **1.75±0.23** | **0** | 9.63%±3.36 | 1.13±0.17 | 3 | 5.89%±6.54 | 1.20±0.17 | 2 | 14.24%±2.92 | 1.03±0.19 | 4 | 16.98%±8.34 | 0.96±0.12 | 4 |
|  | S2 | **1.50%±1.72** | **1.47±0.06** | 1 | 16.77%±6.15 | 0.97±0.33 | 4 | 15.88%±8.16 | 0.96±0.10 | 4 | 19.95%±2.82 | 0.77±0.27 | 6 | 18.32%±1.92 | 0.93±0.17 | 5 |
|  | S3 | **1.97%±1.82** | **1.50±0.21** | 1 | 16.84%±1.66 | 0.96±0.09 | 4 | 17.56%±6.55 | 1.11±0.24 | 3 | 26.23%±4.87 | 1.00±0.18 | 7 | 22.78%±11.99 | 0.80±6.54 | 4 |

Table V
EFFICIENCY METRICS

|  |  | Env1 | | | Env2 | | | Env3 | | | | Env4 | | |
|---|---|---|---|---|---|---|---|---|---|---|---|---|---|---|
|  |  | S1 | S2 | S3 | S1 | S2 | S3 | S1 | S* | S2 | S3 | S1 | S2 | S3 |
| Our method (CPC & CNC) | Vel | **0.91±0.01** | **0.92±0.04** | **0.91±0.01** | **0.88±0.01** | **0.91±0.02** | **0.89±0.02** | **0.93±0.01** | **0.88±0.02** | **0.93±0.01** | **0.92±0.02** | 0.87±0.02 | **0.89±0.01** | **0.90±0.01** |
|  | Length | 39.62±0.30 | 40.05±1.33 | 40.16±0.40 | 30.97±0.69 | 35.00±0.96 | 31.40±0.77 | 39.45±0.39 | **39.59±0.48** | 39.88±0.77 | 40.97±1.39 | 42.37±0.26 | 43.27±0.87 | 41.83±0.78 |
|  | Time | 43.78±0.68 | 43.33±1.41 | 43.90±0.02 | 35.07±0.53 | 31.77±0.14 | 35.31±0.89 | **42.59±0.43** | **45.29±1.57** | **42.57±0.52** | 44.72±2.27 | 48.87±1.29 | **48.47±0.52** | 46.53±0.74 |
| Adding CPC | Vel | 0.85±0.03 | 0.84±0.05 | 0.83±0.02 | 0.86±0.04 | 0.83±0.02 | 0.86±0.11 | 0.90±0.02 | 0.81±0.07 | 0.91±0.02 | 0.91±0.02 | **0.88±0.02** | 0.85±0.02 | 0.85±0.02 |
|  | Length | 40.76±0.56 | 40.31±1.12 | 41.07±1.03 | 28.67±0.91 | 31.18±1.79 | 31.56±1.58 | 40.55±0.97 | 45.52±4.48 | 40.45±0.31 | 39.90±0.40 | 40.13±2.94 | 46.27±4.01 | 43.03±1.22 |
|  | Time | 47.99±1.40 | 47.81±3.80 | 49.91±1.74 | 33.33±2.44 | 37.61±3.19 | 37.52±7.47 | 45.08±1.95 | 56.23±3.72 | 44.29±1.24 | **43.99±1.32** | **45.86±4.07** | 54.24±4.41 | 50.46±2.60 |
| Adding CNC | Vel | 0.86±0.02 | 0.82±0.04 | 0.86±0.03 | 0.84±0.06 | 0.88±0.03 | 0.83±0.10 | 0.89±0.05 | 0.90±0.03 | **0.88±0.02** | 0.83±0.06 | 0.90±0.03 | 0.86±0.02 | 0.86±0.04 |
|  | Length | 38.25±1.00 | 35.85±1.38 | 37.58±3.37 | 32.07±1.01 | 29.75±1.97 | 27.59±1.36 | 38.02±0.84 | 50.92±5.01 | 39.00±1.00 | 40.35±2.20 | 42.62±0.71 | 44.43±4.29 | 42.61±1.86 |
|  | Time | 44.43±1.38 | 43.72±2.83 | 44.72±6.65 | 36.62±2.35 | 35.97±3.73 | 31.23±3.56 | 42.17±1.28 | 55.63±3.43 | 43.03±1.75 | 48.49±3.71 | 52.24±1.94 | 53.51±7.01 | 49.52±3.24 |
| Risk-RRT | Vel | 0.88±0.05 | 0.8±0.04 | 0.86±0.05 | 0.86±0.09 | **0.89±0.01** | 0.88±0.03 | 0.91±0.02 | 0.86±0.04 | 0.85±0.02 | 0.84±0.05 | 0.84±0.05 | 0.85±0.09 | 0.86±0.03 |
|  | Length | 37.14±1.51 | 37.46±1.63 | 36.51±1.04 | **27.97±1.06** | 28.86±0.84 | **27.30±0.23** | 36.56±0.28 | 44.17±2.90 | 38.59±2.00 | 38.28±2.79 | 41.37±2.66 | 44.85±4.64 | 41.46±2.85 |
|  | Time | 42.17±2.99 | 46.32±2.26 | 42.15±2.94 | 33.02±4.07 | 33.70±2.96 | **30.62±0.42** | 40.34±0.74 | 51.51±3.40 | 45.40±2.88 | 45.86±5.13 | 49.43±5.59 | 53.16±6.91 | 48.46±4.73 |
| Bi-Risk-RRT | Vel | 0.86±0.04 | 0.85±0.02 | 0.86±0.03 | 0.85±0.05 | 0.87±0.04 | 0.89±0.03 | 0.91±0.01 | 0.89±0.03 | 0.89±0.03 | 0.88±0.05 | 0.84±0.07 | 0.87±0.05 | 0.81±0.05 |
|  | Length | **35.38±0.81** | **35.23±1.30** | **34.52±0.64** | 28.03±1.95 | **27.61±0.65** | 29.75±3.31 | **36.10±0.41** | 44.88±2.82 | **37.93±1.60** | **38.11±2.24** | 41.96±3.76 | **40.93±2.61** | **40.18±0.73** |
|  | Time | **40.97±1.26** | **41.25±2.15** | **40.75±1.88** | **31.85±2.41** | **30.21±0.82** | 33.52±4.18 | 39.64±1.25 | 54.86±2.59 | 42.65±1.56 | 45.15±0.84 | 47.15±3.70 | 50.99±4.92 | 50.00±0.05 |

Table VI
METRICS OF COMPARATIVE EXPERIMENTS

|  | Ours | | | [15] | | | [6] | | | [8] | | |
|---|---|---|---|---|---|---|---|---|---|---|---|---|
|  | RMSE | TD | Vel | RMSE | TD | Vel | RMSE | TD | Vel | RMSE | TD | Vel |
| Env1(S1) | **0.34±0.01** | **0** | 0.92±0.01 | 0.35±0.02 | 3.65%±0.06 | 0.77±0.11 | 0.41±0.09 | 5.98%±0.10 | 0.90±0.04 | 0.37±0.13 | 14.72%±0.08 | 0.66±0.06 |
| Env2(S2) | **0.15±0.01** | **0.91%±0.81** | 0.92±0.01 | 0.18±0.08 | 6.53%±0.08 | 0.81±0.07 | 0.20±0.02 | 13.5%±0.08 | 0.87±0.05 | 0.27±0.06 | 16.00%±0.06 | 0.70±0.12 |
| Env3(S1) | **0.20±0.01** | **0.12%±0.22** | 0.93±0.01 | 0.21±0.01 | 6.43%±0.02 | 0.81±0.05 | 0.23±0.02 | 23.09%±0.02 | 0.92±0.01 | 0.23±0.02 | 15.52%±0.09 | 0.78±0.04 |
| Env4(S3) | **0.30±0.02** | **1.96%±1.55** | 0.91±0.01 | 0.31±0.01 | 9.56%±0.02 | 0.87±0.04 | 0.39±0.06 | 7.31%±0.03 | 0.80±0.10 | 0.32±0.05 | 11.00%±0.06 | 0.69±0.11 |

## C. Human Awareness Evaluation

In terms of human awareness, we recorded the distances ($D$) between the robot and pedestrians during the navigation process. To quantify human comfort, $D=1.5m$ is considered as the comfort threshold [17]. Besides, the ratio of the duration ($TD$), the minimum distance ($MD$) between pedestrians and the robot during the running process, and the average times of the distance falling below the comfort threshold ($NT$) are recorded to further assess human awareness. The equation of $TD$ is: $TD=t_{inv}/t_{all}$. $t_{inv}$ represents the duration for which the distance is less than the comfort threshold, and $t_{all}$ means the navigation time. The metrics are recorded in Table IV.

Our method achieves the best performance in terms of $MD$ among other methods. As the human number increases, like **S2** and **S3** scenarios in each environment, even though the proposed method enables the $MD$ value to be lower than the comfort threshold, the $NT$ and $TD$ of our methods are the lowest among comparison. Although the CNC algorithm can keep the robot away from the crowd, because the robot does not have the accurate position, its performance is worse than CPC.

## D. Efficiency Evaluation

According to the efficiency of robot navigation tasks, considering the robot running smoothly at high speed is an important factor for navigation, we recorded the average velocity ($Vel$) of the robot to evaluate the efficiency. Additionally, trajectory length and running time are recorded for a comprehensive assessment, which is shown in Table V. Although the focus of this study is accurate localization and human comfort in large-scale and crowded environments, our planner can obtain comparable efficiency to other algorithms.

## E. Analysis of Runtime Performance

To show the results in more detail, we display the experimental result in **Env4 S2**. The trajectory results are shown in Fig. 4, which are obtained in a large-scale and crowded environment with 27 humans. In this scenario, there are a few landmarks in the grey area of the environment, and most humans congregate in the white area. We graphically show the trace of the covariance matrix in Fig. 5(a), and the velocity of the robot in Fig. 5(b)) during the navigation process of different algorithms.

As for the perspective of localization, our method (red line) and the planner that contains CPC (green line) can drive the robot close to the landmarks when the robot is accurate localization. However, the other three methods drive the robot to travel to open areas without enough landmarks for localization. Therefore, as shown in Fig. 5(a), compared with other methods (blue, yellow, and pink lines), the covariance matrix's trace of our algorithm (red line) and the planner that contains CPC (green line) converge quickly. It means that the proposed method and the planner with CPC realize lower pose estimation uncertainty than the comparison.

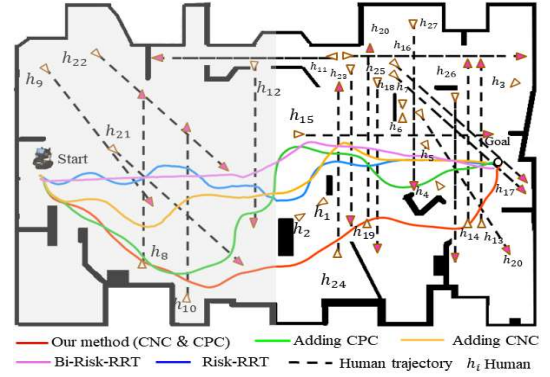

Figure 4. The comparative experiments in **Env4 (S2)**. Humans are shown at their final (starting) positions in the pink (white) triangle whose vertex angle also shows the direction of pedestrians. The grey area is the sparse landmark area, and the white area is the crowded area. The solid black objects are landmarks including walls and static obstacles.

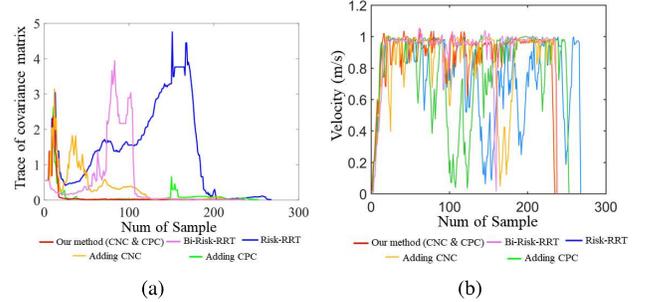

Figure 5. Experimental performance in Env 4. (a) Trace of the covariance matrix of different methods during the navigation process. (b) The velocity of different methods during the navigation process.

From the perspective of the crowd, as shown on the right side of Fig. 4, compared with the other algorithms, our approach can generate the trajectory bypassing the crowded area. The other methods drive the robot into the crowd, which has to make a detour or stop to avoid humans. Because of rushing into a crowded area, other methods are unable to keep a proper distance from humans, which is indicated by the $TD$ and $MD$ in Table IV. Furthermore, by avoiding entering crowds, the proposed method can maintain higher and more stable velocities compared with methods that drive the robot into crowds. In short, this experiment demonstrates the effectiveness of our curiosity-based method in large-scale and crowded environments.

*F. Comprehensive Comparison*

To further verify the performance of the proposed method, we compared it with [15], which considers localization and collision risk, [6], which considers safety and human comfort, and [8], which considers the social navigation behaviors. We randomly selected one scene in each environment for the comparison experiment. We record $RMSE$, Ratio of the Duration $TD$, and Velocity $Vel$ to evaluate the performance.

As shown in Table VI, [15] without considering the crowd gathered to drive the robot enters the crowd. The dynamic humans have a negative impact on the robot's landmark observation and human comfort. [6] and [8] without considering the localization uncertainty do not have an accurate localization in a dense crowd. Accurate localization is the basis of avoiding collisions, which makes the performance of [6] and [8] worse. The proposed method considering crowds and localization uncertainty has the best performance.

## VI. DISCUSSION AND CONCLUSION

Our experiments demonstrate the efficacy of our approach over several baseline methods. The proposed CPC of information-rich areas yields an attraction from a landmark area when the robot's localization is inaccurate, which can result in lower RMSE, TCM, and NM. Besides, CNC in crowded areas can prevent the robot from being blocked by the crowd, which causes a lower TD, and higher MD and NT. Even though crossing the crowd is an optimal choice in terms of trajectory length, our method prevents rushing into the crowd and avoids frequent updates to the trajectory of complex and dynamic crowds. Therefore, our algorithm's velocity curves are higher and smoother, causing a compared time consumption. Besides, since the amount of free space is limited and always changing, the other baseline methods struggle to find collision-free paths in highly dynamic crowded environments, let alone maintain human comfort. On the contrary, our method ensures human comfort while reducing the collision risk. However, this paper currently does not consider the case that humans block out landmarks. In the future, we will add the crowd walking near landmarks as a negative gain into CPC, and encourage the robot to choose the landmark area with less occlusion. In addition, we will conduct physical experiments and establish the benchmark dataset.

In conclusion, our curiosity-based method achieves a human-friendly path with the consideration of localization uncertainty and the distribution of the crowd. Experiments on different scenarios reveal that our method performs well in large-scale and crowded environments.

## ACKNOWLEDGMENT

The authors are grateful to Prof. Jen Jen Chung (Autonomous Systems Lab, ETH Zurich) for her help with the revision.

## REFERENCES

[1] Lukas Huber, Jean-Jacques Slotine, and Aude Billard. Avoiding dense and dynamic obstacles in enclosed spaces: Application to moving in crowds. *IEEE Transactions on Robotics*, 2022.
[2] Adam Bry and Nicholas Roy. Rapidly-exploring random belief trees for motion planning under uncertainty. In *2011 IEEE international conference on robotics and automation*, pages 723–730. IEEE, 2011.
[3] Adarsh Jagan Sathyamoorthy, Utsav Patel, Tianrui Guan, and Dinesh Manocha. Frozone: Freezing-free, pedestrian-friendly navigation in human crowds. *IEEE Robotics and Automation Letters*, 5(3):4352–4359, 2020.
[4] Angeliki Zacharaki, Ioannis Kostavelis, Antonios Gasteratos, and Ioannis Dokas. Safety bounds in human robot interaction: a survey. *Safety science*, 127:104667, 2020.
[5] Michael Everett, Yu Fan Chen, and Jonathan P How. Collision avoidance in pedestrian-rich environments with deep reinforcement learning. *IEEE Access*, 9:10357–10377, 2021.
[6] Kuanqi Cai, Weinan Chen, Chaoqun Wang, Shuang Song, and Max Q-H Meng. Human-aware path planning with improved virtual doppler method in highly dynamic environments. *IEEE Transactions on Automation Science and Engineering*, 2022.
[7] Aiswarya L and Abhra Roy Chowdhury. Human aware robot motion planning using rrt algorithm in industry4.0 environment. In *2021 IEEE International Conference on Intelligence and Safety for Robotics (ISR)*, pages 351–358, 2021.
[8] Vishnu K. Narayanan, Takahiro Miyashita, and Norihiro Hagita. Formalizing a transient-goal driven approach for pedestrian-aware robot navigation. In *2018 27th IEEE International Symposium on Robot and Human Interactive Communication (RO-MAN)*, pages 862–867, 2018.
[9] Keyvan Majd, Shakiba Yaghoubi, Tomoya Yamaguchi, Bardh Hoxha, Danil Prokhorov, and Georgios Fainekos. Safe navigation in human occupied environments using sampling and control barrier functions. In *2021 IEEE/RSJ International Conference on Intelligent Robots and Systems (IROS)*, pages 5794–5800, 2021.
[10] Zichao Zhang and Davide Scaramuzza. Perception-aware receding horizon navigation for mavs. In *2018 IEEE International Conference on Robotics and Automation (ICRA)*, pages 2534–2541. IEEE, 2018.
[11] Tyler Summers. Distributionally robust sampling-based motion planning under uncertainty. In *2018 IEEE/RSJ International Conference on Intelligent Robots and Systems (IROS)*, pages 6518–6523. IEEE, 2018.
[12] Xiaolong Zhang, Yu Huang, Youmin Rong, Gen Li, Hui Wang, and Chao Liu. Optimal trajectory planning for wheeled mobile robots under localization uncertainty and energy efficiency constraints. *Sensors*, 21(2):335, 2021.
[13] Romain Pepy and Alain Lambert. Safe path planning in an uncertain-configuration space using rrt. In *2006 IEEE/RSJ International Conference on Intelligent Robots and Systems*, pages 5376–5381, 2006.
[14] Billy Pik Lik Lau, Brandon Jin Yang Ong, Leonard Kin Yung Loh, Ran Liu, Chau Yuen, Gim Song Soh, and U-Xuan Tan. Multi-agv's temporal memory-based rrt exploration in unknown environment. *IEEE Robotics and Automation Letters*, 7(4):9256–9263, 2022.
[15] Kuanqi Cai, Chaoqun Wang, Shuang Song, Haoyao Chen, and Max Q-H Meng. Risk-aware path planning under uncertainty in dynamic environments. *Journal of Intelligent & Robotic Systems*, 101(3):1–15, 2021.
[16] Han Ma, Jianbang Liu, Fei Meng, Jin Pan, Jiankun Wang, and Max Q-H Meng. A nonuniform sampling strategy for path planning using heuristic-based certificate set. In *2021 IEEE International Conference on Robotics and Biomimetics (ROBIO)*, pages 1359–1366. IEEE, 2021.
[17] Gerald L Stone and Cathy J Morden. Effect of distance on verbal productivity. *Journal of Counseling Psychology*, 23(5):486, 1976.
[18] Jamie Snape, Jur Van Den Berg, Stephen J Guy, and Dinesh Manocha. The hybrid reciprocal velocity obstacle. *IEEE Transactions on Robotics*, 27(4):696–706, 2011.
[19] Chiara Fulgenzi, Anne Spalanzani, Christian Laugier, and Christopher Tay. Risk based motion planning and navigation in uncertain dynamic environment. 2010.
[20] Han Ma, Fei Meng, Chengwei Ye, Jiankun Wang, and Max Q-H Meng. Bi-risk-rrt based efficient motion planning for mobile robots. *IEEE Transactions on Intelligent Vehicles*, 2022.
[21] Pieter M Blok, Koen van Boheemen, Frits K van Evert, Joris IJsselmuiden, and Gook-Hwan Kim. Robot navigation in orchards with localization based on particle filter and kalman filter. *Computers and Electronics in Agriculture*, 157:261–269, 2019.